\pdfoutput=1
\documentclass[letterpaper, 10 pt, conference]{ieeeconf}  

\IEEEoverridecommandlockouts                              

\usepackage[backend=biber,
            url=false,
            isbn=false,
            doi=false,
            backref=false,
            style=ieee,
            natbib=true,
            mincitenames=1,
            maxcitenames=1,
            citestyle=numeric-comp,
            sorting=nyt,
            block=none]{biblatex}
        
\renewcommand{\bibfont}{\small}
\usepackage{subcaption}

\usepackage{amsmath, amssymb, amscd}
\usepackage{gensymb} 

\usepackage{array} 
\usepackage{tabularx}
\usepackage{multirow}
\usepackage{multicol}
\usepackage{booktabs}
\usepackage{tabulary}



\usepackage[colorinlistoftodos]{todonotes}

\title{\LARGE \bf
Kit-Net: Self-Supervised Learning to Kit \\Novel 3D Objects into Novel 3D Cavities
}

\author{Shivin Devgon$^{1}$, Jeffrey Ichnowski$^{1}$, Michael Danielczuk$^{1}$, Daniel S. Brown$^{1}$, Ashwin Balakrishna$^{1}$, \\
Shirin Joshi$^{2}$, Eduardo M. C. Rocha$^{2}$, Eugen Solowjow$^{2}$, Ken Goldberg$^{1}$ 
\thanks{$^{1}$The AUTOLAB at the University of California, Berkeley. 
$^{2}$Siemens Research Lab, Berkeley, CA.
{\tt\small
  \{shivin302, jeffi, mdanielczuk, dsbrown, ashwin\_balakrishna\}\allowbreak @berkeley.edu,
  \{shirin.joshi, eduardo.moura\_cirilo\_rocha,
  eugen.solowjow\}\allowbreak @siemens.com,
  goldberg@berkeley.edu }
}%
}

\begin{document}

\maketitle

\begin{abstract}
In industrial part kitting, 3D objects are inserted into cavities for transportation or subsequent assembly. Kitting is a critical step 
as it can decrease downstream processing and handling times and enable lower storage and shipping costs.
We present Kit-Net, a framework for kitting previously unseen 3D objects into cavities given depth images of both the target cavity and an object held by a gripper in an unknown initial orientation. Kit-Net uses self-supervised deep learning and data-augmentation to train a convolutional neural network (CNN) to robustly estimate 3D rotations between objects and matching concave or convex cavities using a large training dataset of simulated depth images pairs. Kit-Net then uses the trained CNN to implement a controller to orient and position novel objects for insertion into novel prismatic and conformal 3D cavities. Experiments in simulation suggest that Kit-Net can orient objects to have a 98.9\,\% average intersection volume between the object mesh and that of the target cavity. Physical experiments with industrial objects succeed in 18 \% of
trials using a baseline method and in 63\,\% of trials with Kit-Net.
Video, code, and data are available at \url{https://github.com/BerkeleyAutomation/Kit-Net}.
\end{abstract}

\section{Introduction}
\label{sec:introduction}
Kitting is the critical part of industrial automation in order to organize and place 3D parts into complementary cavities. This process saves time on the manufacturing line and frees up space to reduce shipping and storage cost. Automating kitting requires picking and re-orienting a part to a desired position and orientation, and then inserting it into a cavity that loosely conforms to the object geometry. However, automating this process is a great challenge, and in industry, most kitting 
is performed manually.

\begin{figure}[t]
  \centering
  \includegraphics[width=\columnwidth,trim=0 0 187 0, clip]{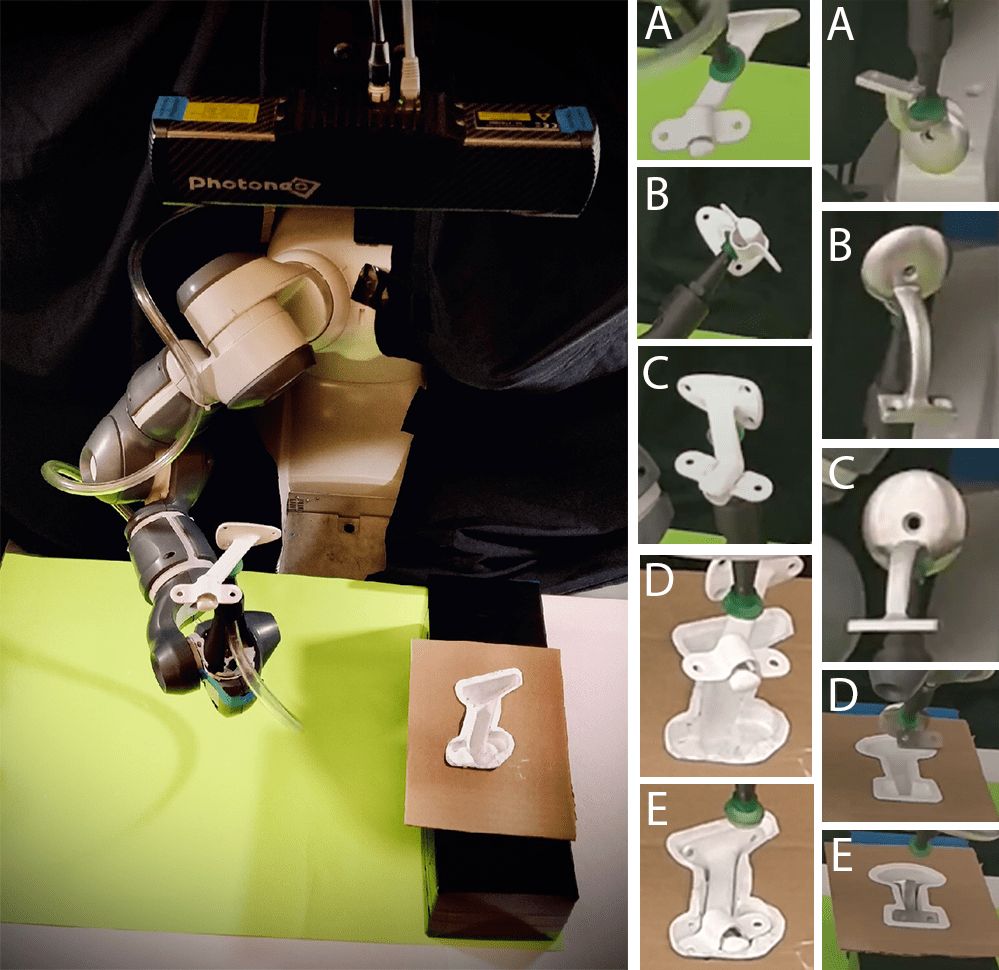}
  \caption{\textbf{Physical experiments using an ABB YuMi robot and a Photoneo depth camera}. Left: A suction gripper holds an object unseen during training time
  near the kitting cavity. 
  Kit-Net orients the object
  for insertion into the cavity through 5 steps. A) Starting state. B) Flip the object 180\degree{} to face the camera and minimize occlusion from the gripper. C) Iteratively orient the object into a goal configuration. 
  D) Flip back and align centroids of the object and cavity to prepare for insertion. E) Insert and release.}
  \label{fig:splash}
\end{figure}

Given a 3D CAD model of the object to be inserted and the desired object pose, one approach is to directly estimate the object pose and the transformation to the desired pose~\cite{ChoyDeepGlobal, xiang2017posecnn}. However, CAD models may not be available for all objects to be kit and are time consuming to create for every object, motivating an algorithm that can kit previously unseen objects without requiring such models. Prior work has considered kitting objects without models, 
but has focused on $SE(2)$ transforms (rotation and translation in a plane) for extruded 2D polygonal objects~\cite{Zakka2020Form2FitLS,zeng2020transporter}. In this work, we propose a method for kitting objects with complex 3D geometries, for which 3D transforms may be required for reliable kitting.


We formalize the problem of rotating and translating a novel 3D object to insert it into a novel kitting cavity and present \emph{Kit-Net}.
Kit-Net is a framework for inserting previously unseen 3D objects with unknown geometry into a novel target cavity given depth images of the object in its current orientation and a depth image of either a flipped (convex) or standard (concave) target cavity. Kit-Net extends prior work from~\citet{CASE_Orienting}, which used simulation and self-supervision to train a deep neural network to directly estimate 3D transformations between the two depth images. Given the trained deep neural network, a depth image of a previously unseen insertion cavity, and a depth image of a previously unseen object, Kit-Net iteratively estimates the $SE(3)$ transform to reorient and insert the object, without requiring detailed knowledge of its geometry. Kit-Net improves on prior work by (a) introducing dataset augmentations that make the controller more robust, (b) using a suction cup gripper to minimize object occlusion during rotation, (c) incorporating 3D translations, and (d) applying the resulting controller to kit novel objects into novel cavities on a physical robot. We evaluate Kit-Net both in simulation and in physical experiments on an ABB YuMi robot with a suction gripper and overhead depth camera. Experiments in simulation suggest that Kit-Net can orient objects to have a 98.9\,\% average intersection volume between the object mesh and that of the target cavity. Physical experiments with 3 industrial objects and cavities suggest that Kit-Net can kit objects at a 63\,\% success rate from a diverse set of initial orientations. 


This paper makes the following contributions:
\begin{enumerate}
\item Formulating the problem of iteratively kitting a novel 3D object into a novel 3D cavity.
\item Kit-Net: a self-supervised deep-learning framework for this problem 
\item Simulation experiments suggesting that Kit-Net can reliably orient novel objects for insertion into prismatic cavities. 
\item Physical experiments suggesting that Kit-Net can significantly increase the success rate of 3D kitting into conformal 3D cavities from 18\,\% to 63\,\% over a baseline inspired by Form2Fit~\cite{Zakka2020Form2FitLS} that only considers 2D transformations when kitting. 
\end{enumerate}







\section{Related Work}
\label{sec:related-work}
There has been significant prior work on reorienting objects using geometric algorithms. \citet{goldberg1993orienting} proposes a geometric algorithm that orients polygonal parts with known geometry without requiring sensors. \citet{akella-orienting-uncertainty} extend the work of Goldberg with sensor-based and sensor-less algorithms for orienting objects with known geometry and shape variation. \citet{kumbla2018enabling} propose a method for estimating object pose via computer vision and then reorient the object using active probing. \citet{grasp-gaits} optimize robot finger motions to reorient a known convex object while maintaining grasp stability. \citet{parts-feeder} propose an end-to-end pipeline for bin-picking, regrasping, and kitting known parts into known cavities. In contrast, Kit-Net can reorient objects without prior knowledge of their 3D models.
\citet{delta-pose-est}, \citet{latent-3d-keypoints}, \citet{Wen2020se3TrackNetD6}, and \citet{CASE_Orienting} use data-driven approaches to estimate the relative pose difference between images of an object in different configurations.  
\citet{delta-pose-est} use a Siamese network to estimate the relative pose between two cameras given an RGB image from each camera. \citet{latent-3d-keypoints} propose KeypointNet, a deep-learning approach that learns 3D keypoints by estimating the relative pose between two different RGB images of an object of unknown geometry, but known category. 
\citet{Wen2020se3TrackNetD6} considers an object-tracking task by estimating a change in pose between an RGBD image of the object at the current timestep and a rendering of the object at the previous timestep, but require a known 3D object model. We use the network architecture from \citet{Wen2020se3TrackNetD6} to train Kit-Net, and extend the self-supervised training method and controller from \citet{CASE_Orienting} to kit novel objects into previously unseen cavities. We find that by extending~\citet{CASE_Orienting} to be more robust to object translations and using a suction gripper to reduce occlusions, Kit-Net is able to learn more accurate reorientation controllers.

There has also been significant interest in leveraging ideas in pose estimation for core tasks in industrial automation.
\citet{Litvak2019LearningPE} leverage CAD models and assemble gear-like mechanisms using depth images taken from a camera on a robotic arm's end-effector. \citet{Stevic2020LearningTA} estimate a goal object's pose to perform a shape-assembly task involving inserting objects which conform to a specific shape template into a prismatic cavity. \citet{Zachares2021InterpretingCI} combines vision and tactile
sensorimotor traces for an object-fitting task involving known holes and object types.
\citet{Huang2020Generative3P} consider the problem of assembling a 3D shape composed of several different parts. This method assumes known part geometry and develops an algorithm to generate the 6-DOF poses that will rearrange the parts to assemble the desired 3D shape. In contrast to these works, we focus on the problem of designing a controller which can reorient and place a novel object within a previously unseen cavity for industrial kitting tasks. 

Object kitting has also seen recent interest from the robotics community. \citet{Zakka2020Form2FitLS} introduce Form2Fit, an algorithm that learns $SE(2)$ transforms to perform pick-and-place for kitting planar objects. In contrast, we consider 6-DOF transforms of 3D objects.
\citet{zeng2020transporter} propose a network for selecting suction grasps and grasp-conditioned placement, which can generalize to multiple robotic manipulation tasks, including pick-and-place for novel flat objects. \citeauthor{zeng2020transporter} focuses on $SE(2)$ rotations and translations for pick-and-place tasks involving novel flat extruded 2D objects. \citeauthor{zeng2020transporter} also presents an algorithm for $SE(3)$ pick-and-place tasks, but only evaluate the algorithm on extruded 2D objects. In contrast, we use Kit-Net to kit novel 3D objects with complex geometries.
\section{Problem Statement}
\label{sec:problem-statement}
\begin{figure}[t]
  \centering
  \vspace{8pt}
  \includegraphics[width=121pt,trim=440 0 440 0, clip]{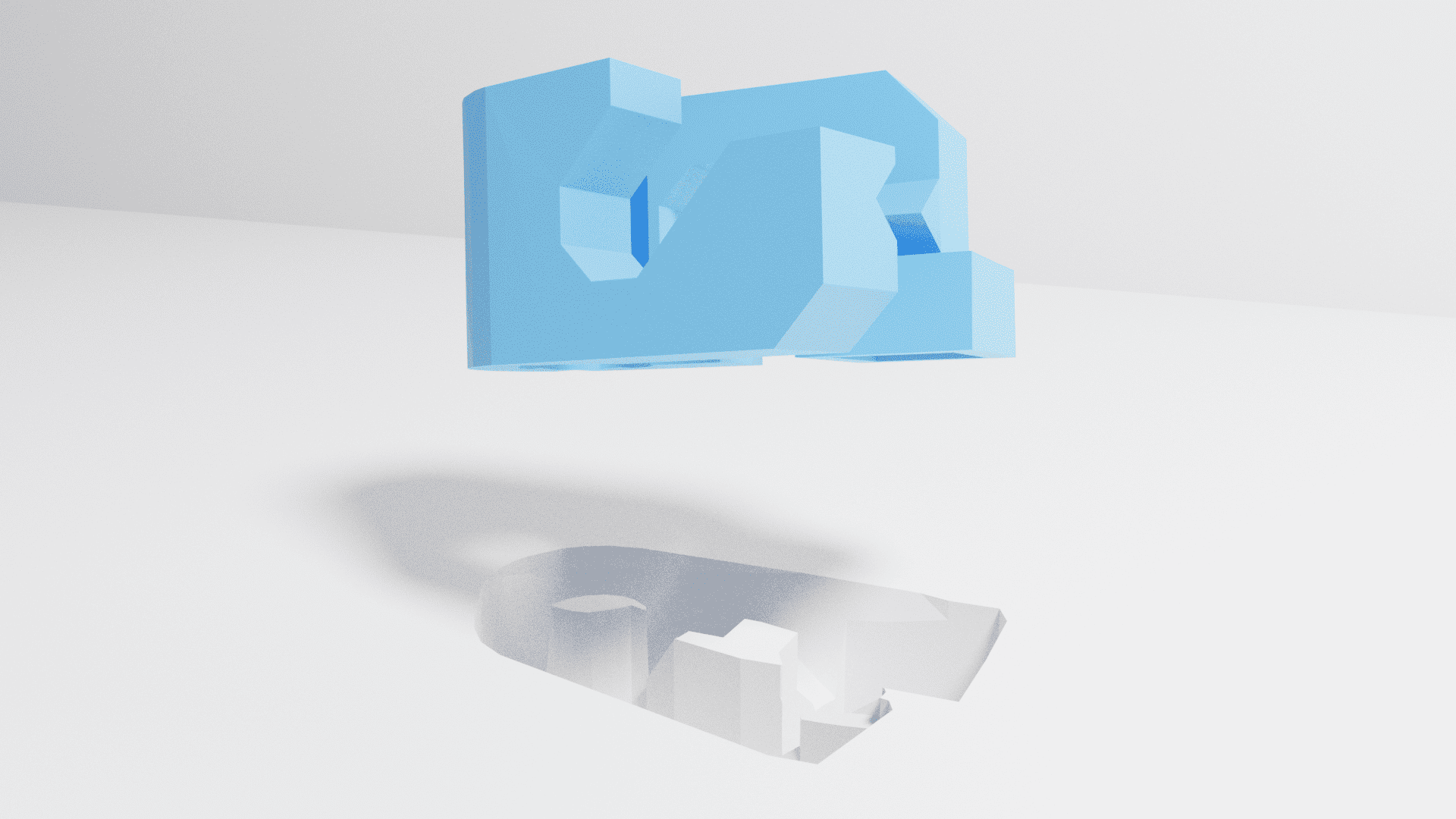}\hfill%
  \includegraphics[width=121pt,trim=440 0 440 0, clip]{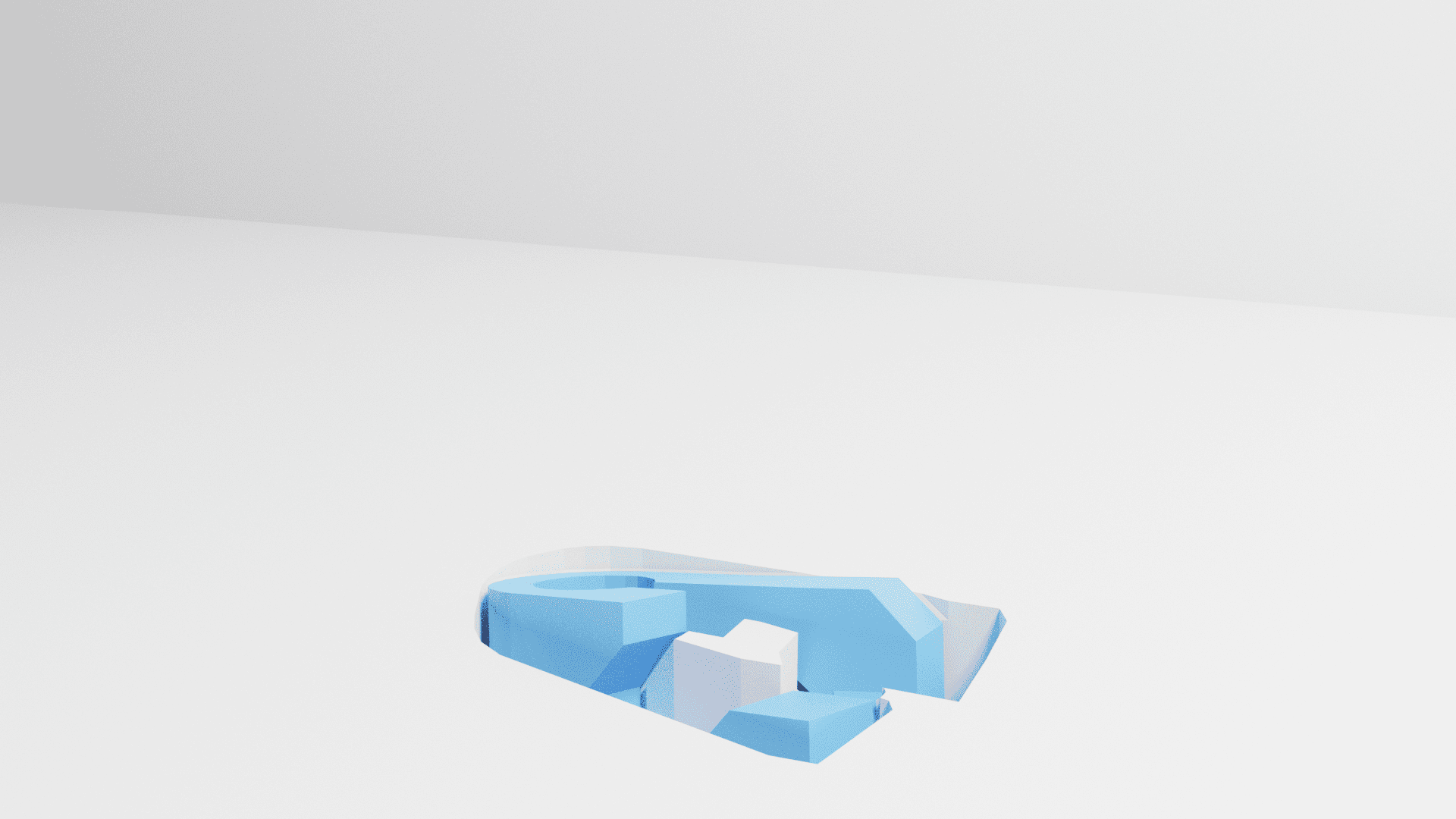}
  \caption{\textbf{Successful Kitting: }Visualization of successfully kitting a 3D object into a concave cavity.}
  \label{fig:mug-cavity-2}
\end{figure}

Let $T^s \in SE(3)$ be the initial 6D pose of a unknown 3D rigid object $O$ in the world coordinate frame, consisting of a rotation $R^s \in SO(3)$ and a translation $t^s \in \mathbb{R}^3$. Given $O$ with starting pose $T^s$ and a kitting cavity $K$, let $\mathcal{G} \subset SE(3)$ be the set of goal 6-DOF poses of object $O$ that result in successful kitting. The goal is to orient object $O$ to $T^g \in \mathcal{G}$, where $T^g$ consists of rotation $R^g$ and translation $t^g$. Figure~\ref{fig:mug-cavity-2} shows a simulated example where a 3D object $O$ is successfully kitted into a concave cavity $K$.

\subsection{Assumptions}
\label{subsec:formulation}
We assume access to depth images of a rigid object $O$ and a kitting cavity $K$. The cavity image may be taken with the cavity either in its standard, concave orientation (i.e., open to object insertion), or flipped, convex orientation (i.e., mirroring the shape of the object to be inserted).
We also assume that orienting $O$ to a pose in $T^g \in \mathcal{G}$ and releasing the gripper results in a successful kitting action.

\subsection{Input}
\label{subsec:input}
Let $I^s \in \mathbb{R}^{H \times W}$ be a depth image observation of the object in initial pose $T^s$, and
$I^k \in \mathbb{R}^{H \times W}$ be the depth image observation of a kitting cavity, $K$. See Figure~\ref{fig:clamshell-cavities} for physical examples of objects and kitting cavities. We use a depth map as input to the neural network only to be consistent with depth-only methods of grasping such as \cite{mahler2017dex} and \cite{danielczuk2020exploratory}.

\subsection{Output}
\label{subsec:output}
The goal is to successfully kit an unknown 3D object $O$ into a novel 3D cavity $K$ (Fig.~\ref{fig:splash}, Fig.~\ref{fig:mug-cavity-2}). Thus, we aim to transform the initial pose $T^s$ into a goal pose that fits into the cavity (i.e., $T^g \in \mathcal{G}$). For objects with symmetries, the objective is to estimate and orient objects relative to a (symmetric) orientation that results in successful insertion into the cavity K.

\section{Kit-Net Framework}
\label{sec:method}
We present Kit-Net, a framework that first reorients the object into a pose that can be successful kit in a desired cavity, and then translates and inserts the object into the cavity. We do this by learning to estimate ${_s}T^g \in SE(3)$, a relative transformation consisting of rotation ${_s}R^g$ and translation ${_s}\hat{t}^g$, which transforms the object from $T^s$ to $T^g$ for some $T^g \in \mathcal{G}$ in the world coordinate frame. The overall approach is to (1) compute an estimate of ${_s}R^g$, denoted ${_s}\hat{R}^g$, given only image observations $I^s$ and $I^k$, (2) iteratively reorient the object according to  ${_s}\hat{R}^g$ and (3) translate the object by ${_s}\hat{t}^g$, an estimate of the translation between the start and goal object translations, such that $\{\hat{R}^g,\hat{t}^g\} \in \mathcal{G}$ and the object can be successfully dropped or inserted into kitting cavity by releasing the gripper.

We first discuss preliminaries (Section~\ref{subsec:prelims}) and then describe the key new ideas in training Kit-Net (Section~\ref{subsec:kit-net-training}) which make it possible to design a controller to rotate and translate an object to fit it in a cavity (Section~\ref{subsec:kit-net-alg}).

\subsection{Preliminaries: Estimating Quaternion Rotations in 3D}
\label{subsec:prelims}
\citet{CASE_Orienting} presented a self-supervised deep-learning method to align
two 3D objects.
The method takes two depth images as input: $I^s$, an image of the object in its current orientation $R^s$; and $I^g$, an image of the same object in its desired goal orientation $R^g$ in the world coordinate frame. It trains a deep neural network $f_\theta:\mathbb{R}^{H\times W}\times\mathbb{R}^{H\times W} \rightarrow SO(3)$
to estimate the rotation ${_s}\hat{R}^g$ (parametrized by a quaternion) in the object coordinate frame between the pair of images $(I^s, I^g)$. In simulation experiments, we use the object's center of mass, while  for physical experiments, we use the robot arm's end effector as the center of the object coordinate frame.
Then, using a proportional controller, it iteratively rotates the object to minimize the estimated rotational difference. This controller applies $\eta{_s}\hat{R}^g$ to the object in the object coordinate frame until
the network predicts that the current object rotation $R^s$ is within a tolerance $\delta$ (e.g., $\delta = 0.5\degree$) of $R^g$, or until the controller reaches an iteration limit. The tunable constant $\eta$ is the Spherical-Linear intERPolation (slerp) factor describing the proportion of $\hat{R}^g$ that the controller will apply to $O$. \citeauthor{CASE_Orienting} use $\eta=0.2$ and an iteration limit of 50 rotations.

For training,~\citeauthor{CASE_Orienting} generate a dataset consisting 200 pairs of synthetic depth images for each of the 698 training objects with random relative rotations, for a total of 139\,600 pairs. To account for parallax effects, each pair of images were generated from a fixed translation relative to the camera.
\citeauthor{CASE_Orienting} propose three loss functions to train $f_\theta$: a cosine loss, a symmetry-resilient loss, and a hybrid of the two, with the hybrid loss outperforming the first two. Note that \citeauthor{CASE_Orienting} do not consider cavity insertion tasks, which is complicated by the need to reason about the translations and required alignment with a cavity. 

Kit-Net improves on \citeauthor{CASE_Orienting} by (1) introducing dataset augmentations to make the controller more robust, (2) using a suction cup gripper to minimize object occlusion during rotation, and (3) incorporating translations into the controller to enable kitting. We discuss these contributions in the following sections.

\subsection{Kit-Net Dataset Generation, Augmentation, and Training}
\label{subsec:kit-net-training}
Kit-Net 
trains a neural network $f_\theta$ with a self-supervised objective by taking as input pairs of depth images $(I^s, I^g)$ and estimating ${_s}\hat{R}^g$ from image pair $(I^s, I^g)$. As in \citeauthor{CASE_Orienting}, $f_\theta$ encodes each depth image into a length 1024 embedding, concatenates the embeddings, and passes the result through two fully connected layers to estimate a quaternion representation of the rotational difference between the object poses. 

\subsubsection{Initial Dataset Generation}
In contrast to \citeauthor{CASE_Orienting}, we are interested in kitting, rather than just reorienting an object in the robot gripper. Thus, in this paper we focus on two types of kitting cavities: prismatic cavities (Fig.~\ref{fig:prism-task}) and conformal cavities (Fig.~\ref{fig:clamshell-cavities}).
We generate a separate dataset for each type of cavity and train a separate network for each dataset. 
To generate both datasets, we use the set of 698 meshes from \citet{mahler2019learning}. For each mesh, we generate 512 depth image pairs, for a total of 357\,376 pairs. To do this we first generate a pair of rotations $(R^s, R^g)$, where $R^s$ is generated by applying one rotation sampled uniformly at random from $SO(3)$ to $O$, and $R^g$ is generated by applying a random rotation with rotation angle less than 30 degrees onto $R^s$. To generate the conformal cavity dataset, we then obtain a pair of depth images $(I^s, I^g)$ by rendering the object in rotations $R^s$ and $R^g$ from an overhead view. The pair is labeled with the ground-truth rotation difference between the images. To generate the prismatic cavity dataset we follow the same process, except we fit and render a prismatic box around the rotated object (Fig.~\ref{fig:prism-task-eval-objects}) and render the depth image pairs (Fig.~\ref{fig:prism-task}). This results in two datasets, each containing 357\,376 total labeled image pairs $(I^s, I^g)$ with ground-truth rotation labels ${_s}R^g$. 

\subsubsection{Data Augmentation}
We found that simply training a network directly on these datasets results in poor generalization to depth images from the physical system which contain (1) sensor noise, (2) object occlusions from the object, the arm, or gripper, and (3) 3D object translations within the image. To address these three points, we introduce dataset augmentations to ease network transfer from simulated to real depth images. 
To simulate noise and occlusion in training, we randomly zero out 1\,\% of the pixels in each depth image and add rectangular cuts of width 30\,\% of zero pixels to the image.
To simulate translations in training,
we randomly translate the object across a range of 10\,cm in the $x$, $y$, and $z$ axes with respect to the camera in the simulated images. We also crop the images at sizes from 5\,\% to 25\,\% greater than the object size with center points offset from the object's centroid by 5 pixels to simulate
$(I^s, I^g)$ pairs generated from objects and cavities outside the direct overhead view. 


\subsubsection{Training} We adopt the network architecture from \citet{Wen2020se3TrackNetD6}, as it is designed to be trained in simulation and demonstrates state-of-the-art performance on object tracking~\cite{DengPoseRBPF} by regressing the relative pose between two images. We use the hybrid quaternion loss proposed by \citet{CASE_Orienting}. The network is trained with the Adam optimizer with learning rate 0.002, decaying by a factor of 0.9 every 5 epochs with an L2 regularization penalty of $10^{-9}$.

\subsection{Kit-Net Suction Gripper}
Kit-Net uses an industrial unicontact suction gripper from \citet{mahler2017suction} to grasp the object for kitting. In contrast, \citeauthor{CASE_Orienting} used a parallel jaw gripper. We find the suction gripper to be better suited for kitting because it reduces gripper occlusions 
and enables the robot to position the object directly inside the kitting cavity.

\subsection{Kit-Net Controller} \label{subsec:kit-net-alg}
The Kit-Net controller consists of two stages: rotation and translation.

\subsubsection{Rotation}
Kit-Net first re-orients an object using the depth image of the current object pose $I^s$ and a depth image of the goal cavity pose $I^g$. In preliminary experiments, we found the rotation parameters used by \citet{CASE_Orienting} (Section~\ref{subsec:prelims}) to be overly conservative. Thus, to speed up the alignment process, we use a larger slerp factor of $\eta=0.8$. If the network predicts a rotation difference of less than $\delta=5\degree$, then we assume that the object is close enough to the required pose for kitting into the cavity and terminate the rotation controller. Because of the larger slerp value, Kit-Net is able to quickly reorient the object, thus we terminate the rotation controller after a maximum of 8 sequential rotations (8 iterations).

\subsubsection{Translation}
We iteratively refine the rotation in $SO(3)$ first before aligning translation to ensure that the translation estimation is as accurate as possible. Once the rotation controller terminates, Kit-Net computes a 2D translation to move the object directly over the target cavity, and then lowers the object and releases it into the cavity. To calculate the 2D translation, we project the final depth image $I^s$ of the object after rotation and the depth image of the cavity $I^g$ to point clouds in the world coordinate frame. We know the position of the object in the gripper and the cavity on the workspace so we can segment out the background and isolate the object and cavity in the point clouds. Then, we perform centroid matching between the two point clouds to estimate the 2D translation.
\section{Simulation Experiments}
\label{sec:experiments}
We first discuss metrics to evaluate performance in Section~\ref{subsec:metrics}. We then introduce a baseline algorithm (Section~\ref{subsec:baseline}) with which to compare Kit-Net and present experimental results in Section~\ref{subsec:results}. In experiments, we first evaluate Kit-Net on re-orienting novel objects with unknown geometry into a target prismatic box in simulation (Section~\ref{subsubsec:prismatic}).
\subsection{Baselines}
\label{subsec:baseline}
\subsubsection{Random Baseline}
We also compare Kit-Net with a baseline that applies a randomly sampled rotation but with the correct rotation angle to evaluate how important precise reorientation is for successful kitting.

\subsubsection{2D Baseline}
To evaluate the importance of estimating 3D rotations for successful kitting, we compare Kit-Net to a baseline inspired by 
Form2Fit~\cite{Zakka2020Form2FitLS}, which only considers 2D rotations when orienting objects for kitting. The baseline (1) aligns the centroids of the point clouds of the object and the cavity and (2) searches over possible rotations about the $z$-axis at a 1\textdegree~discretization to find the rotation that minimizes Chamfer distance between the centroid-aligned point clouds.

\subsection{Metrics}
\label{subsec:metrics}
\subsubsection{Object Eccentricity}
We categorize test objects by their \emph{eccentricity}, which provides a measure of kitting difficulty. This categorization
follows the intuition
that objects that are more elongated along certain dimensions than others have a smaller set of acceptable orientations in which they can be successfully kit into a cavity. Let the eccentricity $\varepsilon$ of a 3D object be $\varepsilon = A - 1$, where $A$ is the aspect ratio (ratio of longest side to shortest side) of the minimum volume bounding box of the object. This definition generalizes the 2D definition of eccentricity from \citet{GoldbergEccentricity2000} to 3D. Under this definition, a sphere has $\varepsilon = 0$, and if one axis is elongated by a factor $p$, then the resulting ellipsoid has eccentricity $p-1$. This definition is also consistent with the intuition provided earlier, as a sphere is entirely rotationally symmetric, and thus does not require any reorientation for kitting. By contrast, the ellipsoid will require reorientation to ensure that its longer side is aligned to a region with sufficient space in the cavity. Thus, we use objects with high eccentricity in evaluating both Kit-Net and the baselines, as these objects pose the greatest challenge for kitting in practice.

\subsection{Results}
\label{subsec:results}
When evaluating Kit-Net in simulation, we have access to ground-truth object and cavity geometry. Thus, we evaluate kitting performance using the following percent fit metric:
\begin{equation}\label{eq:percent-fit}
\hat{\kappa}(I^s, I^g, \hat{{_s}T^g}) = \frac{1}{N} \sum_{i=1}^N \mathbf{1}_{(p_i \in K)},
\end{equation}
where we uniformly sample $N$ points within the object volume at configuration $\hat{{_s}R^g}R^s$ (after the target object has been rotated for insertion) and count the portion of sampled points that also lie within cavity. This metric can be efficiently computed using ray tracing and effectively estimates how much of the object fits inside the target mesh after the predicted rotation.
In experiments, we use $N=10^4$ sampled points to evaluate $\hat{\kappa}$. Assuming the true percent fit metric is $\kappa$, a 95\,\% confidence interval for $\kappa$ is $\hat{\kappa} \pm 1.96\sqrt{\frac{\hat{\kappa}(1-\hat{\kappa})}{10^4}}$. For example, if $\hat{\kappa} = 0.99$, then $\kappa$ lies between $(0.988, 0.992)$ with 95\,\% confidence. 

\subsubsection{Simulated Kitting into a Prismatic Target}
\label{subsubsec:prismatic}
\begin{figure}[t]
  \vspace{8pt}
  \centering
  \includegraphics[width=0.8\linewidth]{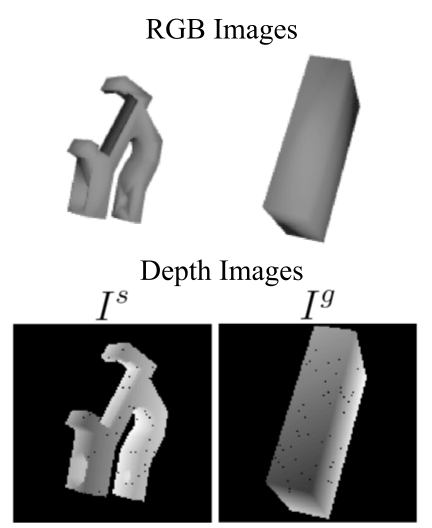}
  \caption{\textbf{Endstop Holder and Target Prismatic Cavity in Simulation: } Given $I^s$, an image of an object in some configuration (top left) and $I^g$, an image of a target prismatic box to which the object must be aligned (top right), the objective is to find a 3D rotation $\hat{{_s}R^g}$ that would allow the object to fit within the box. In simulation experiments, $R^g$ is a $30$\degree~rotation from $R^s$. The image in the figure shows an object that must be rotated by 30\degree~ to perfectly fit it inside the prism. 3D models corresponding to $I^s$ and $I^g$ are shown in the bottom row for clarity. 
  }
  \label{fig:prism-task}
\end{figure}

We first study whether Kit-Net can orient objects into alignment with a prismatic cavity that loosely conforms to their 3D geometry in simulation. Precisely, we first generate the prismatic cavity for the target by creating a mesh with faces corresponding to its minimum volume bounding box. We then rotate both the prismatic cavity and target to random orientations within 30 degrees of each other. The objective is to apply a rotation ${_s}\hat{R}^g$ that will allow the object to fit into the cavity. An example image pair of an object and an associated prismatic cavity is shown in Figure~\ref{fig:prism-task}.

\begin{figure}[t]
  \centering
  \vspace{8pt}
  \includegraphics[width=\linewidth]{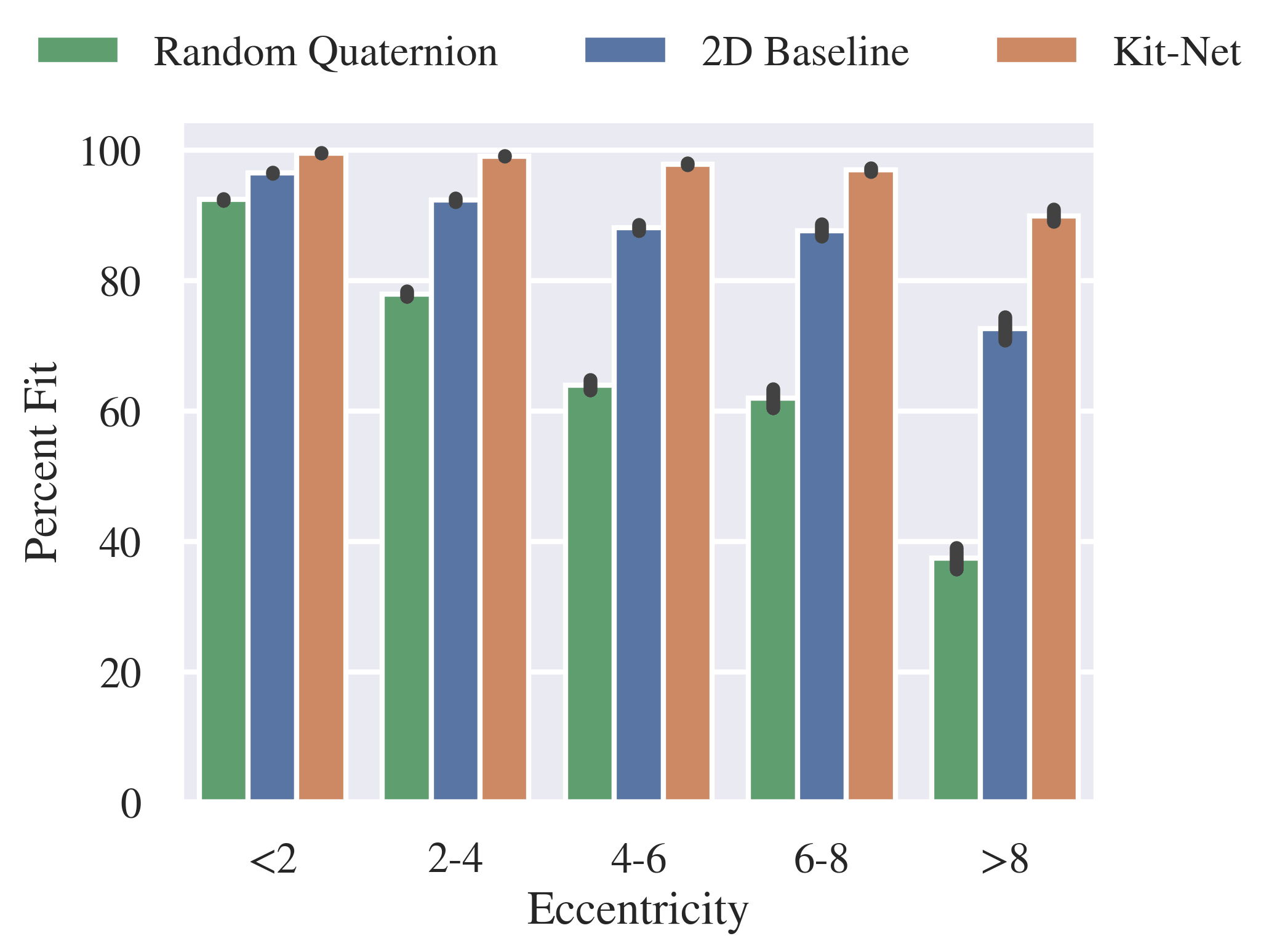}
  \caption{\textbf{Aligning Objects to Prismatic Cavities in Simulation: }We evaluate Kit-Net's ability to align objects with prismatic cavities under the percent fit metric introduced in Section~\ref{subsec:metrics} across 512 depth image pairs for each of 174 objects not seen during training. Given $(I^s,I^g)$, the network predicts $\hat{{_s}R^g}$ that will allow it to fit inside the cavity. We bin results by object eccentricity and observe that the mean percent fit decreases for objects of higher eccentricity. Kit-Net outperforms both the 2D and random baselines by a greater amount as object eccentricity increases.}
  \label{fig:mean-percent-fit-ecc}
\end{figure}

Fig.~\ref{fig:mean-percent-fit-ecc} shows the percent fit across 174 unseen test objects. We use the eccentricity $\epsilon$ of the objects to sort them into 5 bins of increasing difficulty (increasing $\epsilon$). We find that Kit-Net is able to reliably kit novel objects, significantly outperforming the 2D rotation baseline. When averaged across all eccentricities, Kit-Net achieves an average fit of 98.9\,\% compared to an average fit of 93.6\,\% for the 2D baseline and 83.1\,\% when applying a random 30\degree~quaternion. These results demonstrate the need for 3D rotations to solve complex kitting problems. Figure~\ref{fig:mean-percent-fit-ecc} demonstrates that Kit-Net is robust to highly eccentric objects which require the most precision for kitting. Kit-Net achieves an average fit of 89.9\,\% for objects with eccentricity greater than 8. The 2D rotation baseline performs especially poorly for these difficult objects and achieves an average fit of only 72.7\,\% while applying a random 30\degree~quaternion results in an average fit of just 37.4\,\%.

\begin{figure}[t]
  \centering
  \subcaptionbox*{Industrial Part}{\includegraphics[height=50pt, trim=333 48 418 33, clip]{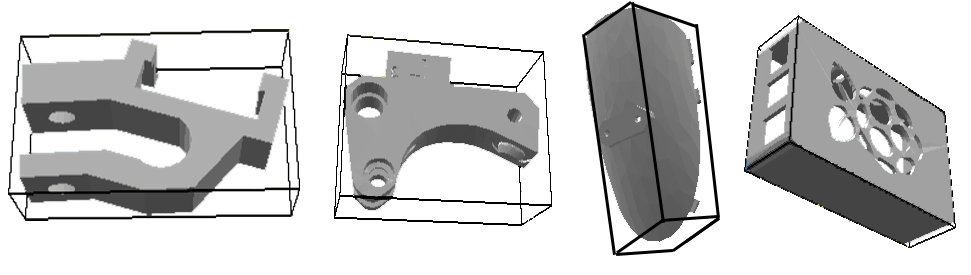}}%
  \hfill%
  \subcaptionbox*{Shield Part}[42pt]{\includegraphics[height=50pt, trim=581 18 250 0, clip]{figures/Industrial_Prismatic_Cavity_Task_Objects.png}}%
  \hfill%
  \subcaptionbox*{Endstop Holder}{\includegraphics[height=50pt, trim=7 53 670 28, clip]{figures/Industrial_Prismatic_Cavity_Task_Objects.png}}%
  \hfill%
  \subcaptionbox*{Raspberry Pi Case}[62pt]{\includegraphics[height=50pt, trim=743 26 14 14, clip]{figures/Industrial_Prismatic_Cavity_Task_Objects.png}}
  \caption{\textbf{Examples of Novel Objects for Kit-Net Simulation Experiments: }
  The four test objects are unseen during training and have eccentricity greater than 2, meaning their minimum volume bounding boxes are narrow and long. An outline of the corresponding minimum volume bounding box is shown around each part.}
  \label{fig:prism-task-eval-objects}
\end{figure}

\begin{figure}[t]
  \centering
  \vspace{8pt}
  \includegraphics[width=0.48\textwidth]{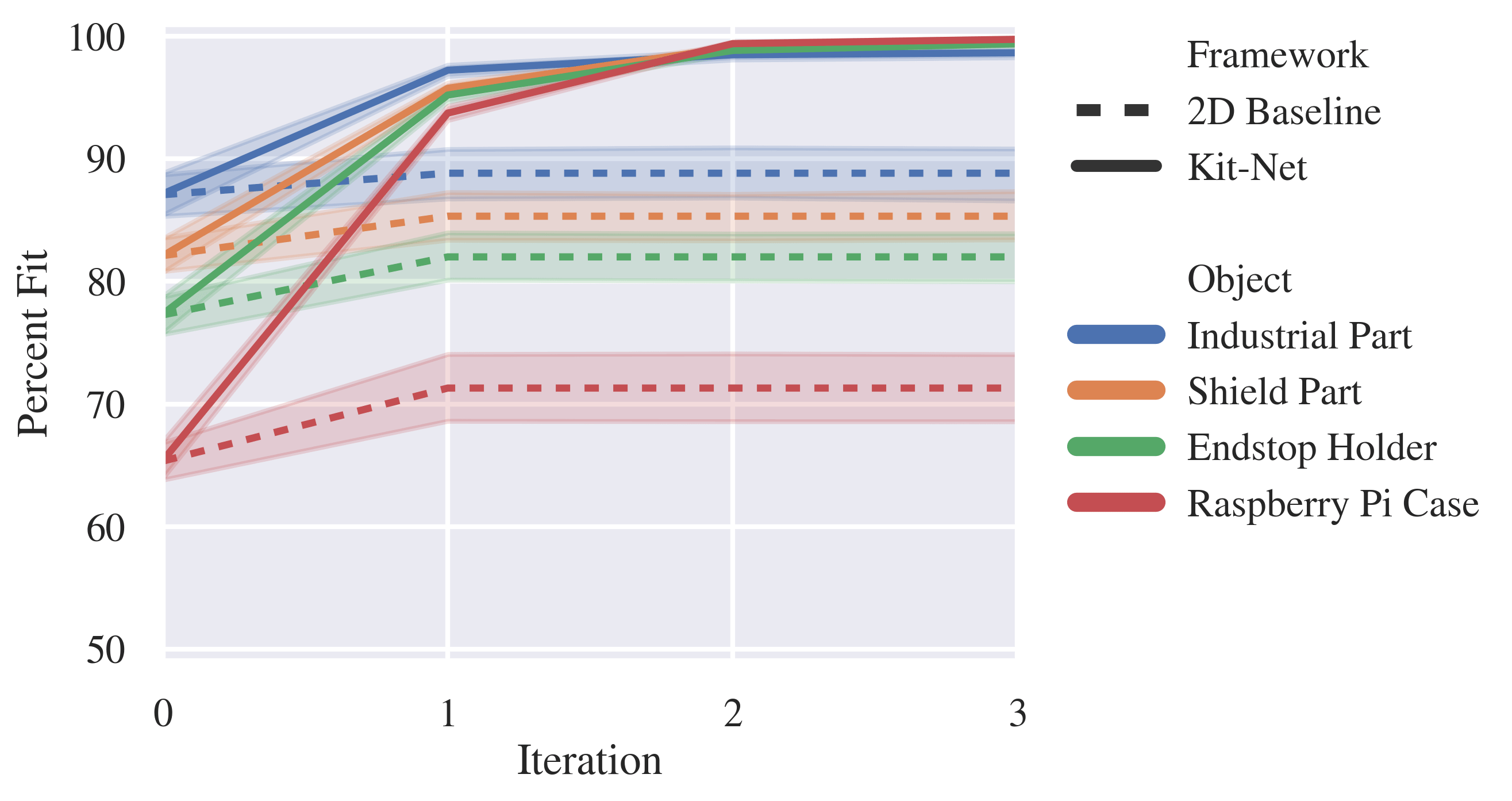}
  \caption{\textbf{Kit-Net Simulation Results: } We visualize data from 100 runs on each of the 4 objects shown in Figure~\ref{fig:prism-task-eval-objects}. All objects require a 30\degree~rotation to be in alignment with the prismatic target at iteration 0, but their initial percent fits differ due to different eccentricities. Results suggest that Kit-Net is able to successfully align all 4 objects with their respective prismatic cavities while the baseline, which restricts itself to 2D rotations, performs significantly worse on all 4 objects.}
  \label{fig:percent-fit-runs}
\end{figure}

As described in Section~\ref{subsec:kit-net-alg}, Kit-Net iteratively orients each object using the controller until $\hat{{_s}R^g}\leq 5\degree$, or until we hit the stopping condition of 8 iterations. Our previous results in Fig.~\ref{fig:mean-percent-fit-ecc} suggests that Kit-Net can consistently align objects within 5 controller steps.
To better visualize the ability of Kit-Net to rapidly reorient an object for kitting, we plot the per-iteration performance of Kit-Net for 4 test objects unseen at training time with high eccentricity ($\epsilon \geq 2$).  Fig.~\ref{fig:prism-task-eval-objects} shows renderings of these objects along with outlines of the corresponding prismatic kitting cavities. Fig.~\ref{fig:percent-fit-runs} shows the average per-iteration percent fit across 100 controller rollouts of randomly sampled $(I^s,I^g)$ pairs for each object. We find that Kit-Net is able to consistently align objects with their target prismatic cavities, and achieves a median fit percentage of 99.4\,\% after only 3 successive iterations of the controller. By contrast, the 2D baseline is not able to surpass an average fit of 90\,\% for any of the objects. The results in Fig.~\ref{fig:percent-fit-runs} demonstrate the importance of iteratively reorienting parts and demonstrates that applying multiple iterations of the rotation output by the trained network can greatly help to reduce the error between $\hat{{_s}R^g}R^s$ and $R^g$ as compared to a single iteration. 

\section{Physical Experiments}
\label{sec:physical-experiments}
Our previous experiments studied the effectiveness of Kit-Net for insertion tasks involving prismatic cavities. However, many physical kitting tasks involve non-prismatic cavities (e.g., Fig.~\ref{fig:clamshell-objects} and Fig.~\ref{fig:clamshell-cavities}). In this section, we study how Kit-Net can be used to kit objects in physical trials using depth images of the types of cavities shown in Fig.~\ref{fig:clamshell-cavities}. We call these \emph{conformal cavities}, as they ``conform'' (to some degree) to the object shape. 

In these experiments, we use a quaternion prediction network trained to predict the quaternion that will rotate a simulated depth image of an object to another simulated depth image of the same object in a different pose. We propose two possible methods for applying this trained network to kitting. Our first method is designed to work well with the clamshell cavities shown in Fig.~\ref{fig:clamshell-cavities}. Rather than image the hole of the cavity, we define a \emph{convex conformal cavity} to be the depth image of the inverted cavity. To obtain these depth images, we flip the cavity so the hole is pointing down and take a depth image of the positive mass of the cavity. The left image in Fig.~\ref{fig:clamshell-cavities} shows examples of these convex conformal cavities. Our second method works with a \emph{concave conformal cavity}, like that shown in Fig.~\ref{fig:mug-cavity-2} and the right image in Fig.~\ref{fig:clamshell-cavities}, that are formed as impressions into a surface. These types of cavities cannot simply be flipped upside down to obtain a depth image of their shape. Instead, we take a depth image of the actual cavity (where the cavity has negative mass) and rotate it 180\degree~about its principal axis.

We discuss the results for applying Kit-Net to novel convex conformal cavities in Section~\ref{subsubsec:real-positive} and to novel concave conformal cavities Section~\ref{subsubsec:real-negative}.
For the physical kitting experiments we measure success using a binary success metric for insertion by visually inspecting whether or not the object is completely contained in the target cavity.

\subsection{Physical Kitting into Convex Conformal Cavities}
\label{subsubsec:real-positive}

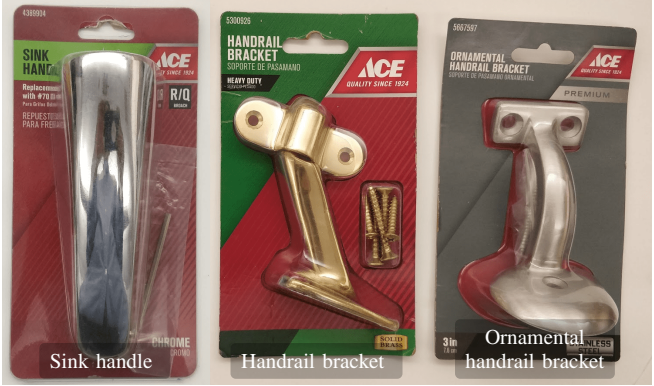
\begin{figure}[t]
  \vspace{8pt}
  \centering
  \begin{tikzpicture}[label/.style={inner sep=4pt, rounded corners=2pt, color=white, fill=black, fill opacity=0.25, text opacity=1, align=center, font=\footnotesize, yshift=2pt}]
    \node [inner sep=0] (img) {%
     \includegraphics[width=\columnwidth]{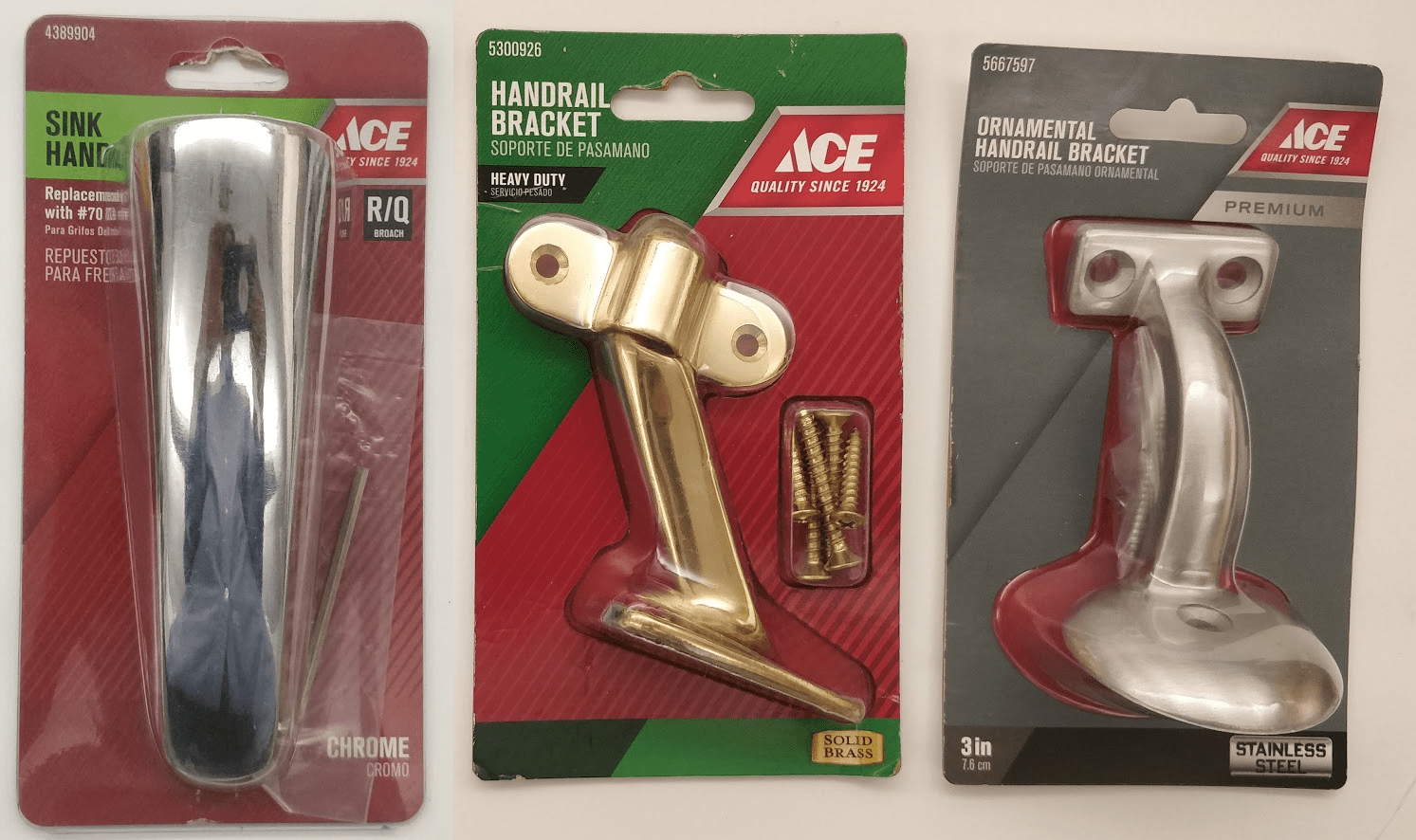}};
    \node [label, anchor=south west, xshift=14pt] at (img.south west) { Sink handle };
    \node [label, anchor=south, xshift=-4pt] at (img.south) { Handrail bracket };
    \node [label, anchor=south east, xshift=-14pt] at (img.south east) { Ornamental \\ handrail bracket };
  \end{tikzpicture}
  \caption{\textbf{Objects for Kit-Net Physical Experiments: }We use 3 packaged industrial objects available in hardware stores. 
   We selected objects for their complex geometries that make precise orientation critical for effective kitting.}
  \label{fig:clamshell-objects}
\end{figure}

\begin{figure}[t]
  \centering
  \includegraphics[width=\columnwidth]{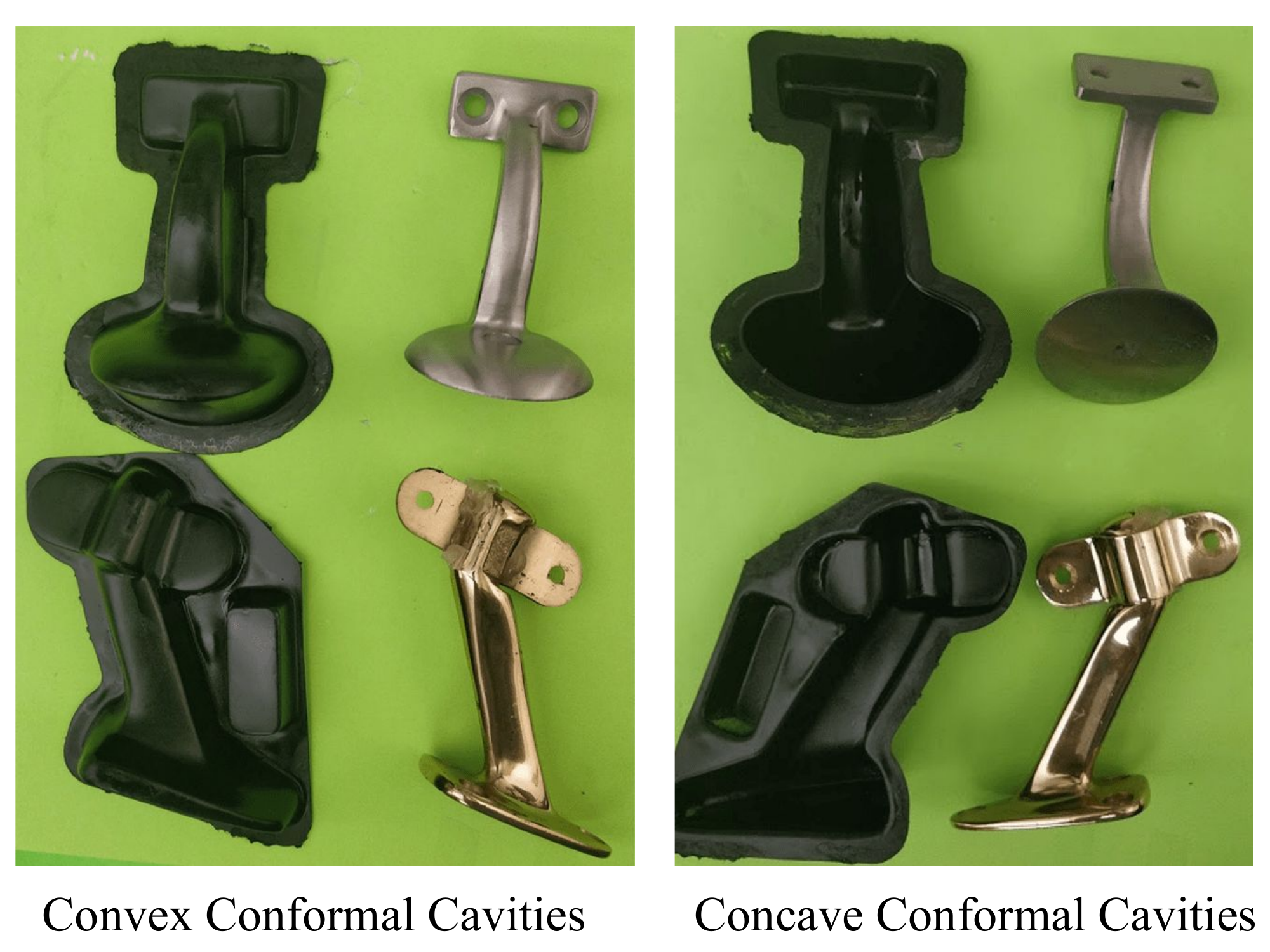}
  \caption{\textbf{Examples of Physical Kitting Cavities: }The handrail bracket (bottom) and the ornamental handrail bracket (top), next to the corresponding convex cavity (left) and concave cavity (right).}
  \label{fig:clamshell-cavities}
\end{figure}

We evaluate Kit-Net in physical kitting trials using an ABB-YuMi robot with a Photoneo depth camera (Fig.~\ref{fig:splash}) on 4 packaged objects available in hardware stores and which are unseen during training (Fig.~\ref{fig:clamshell-objects}). 
To prepare objects for kitting, we extract each object and kitting shell from its packaging and spray paint the shell to facilitate depth sensing (Fig.~\ref{fig:clamshell-cavities}). We then place the kitting cavity open end down directly under the camera and image the cavity to generate $I^g$ before flipping it and moving it onto the workspace to expose its opening for the insertion task. For each trial, we insert the object into the cavity by hand, grasp it using the robot's suction gripper, translate it to be directly under the camera, and apply a random rotation of either $30\degree$ or $60\degree$about a random axis to simulate grasping the object from a bin in a non-uniform pose. Then, the robot flips the object to face the overhead depth camera and have the suction gripper occluded from the camera by the object. This process is illustrated in Fig.~\ref{fig:splash}. 

Kit-Net then orients the object using the learned controller, and matches centroids between the object and cavity for insertion before flipping it again and attempting to kit it. Table~\ref{table:positive-results} shows the number of successful kitting trials (out of 10 per object) of Kit-Net and the 2D baseline across 3 objects. We report a kitting trial as successful if the object is fully contained within the cavity using visual inspection. We observe that Kit-Net outperforms the baseline for 30\degree~initial rotations on 2 of the 3 objects, performing similarly to the baseline on the sink handle. We find that Kit-Net significantly outperforms the baseline on all objects for 60\degree~initial rotations.

The main failure modes are due to errors in the centroid matching procedure, as illustrated in Fig.~\ref{fig:failure-modes}. On the 30 degree sink handle task, Kit-Net aligned it correctly every time, but the centroid matching was off by about 5\,mm, and there is no slack at the top of the cavity.


\begin{table}[h]
\centering
\begin{tabular}{l c c c}\toprule 
 Object & Angle & 2D Baseline & Kit-Net \\
 \midrule
 Handrail bracket & 30\degree & 3/10 & \bf 10/10 \\ 
 Ornamental handrail bracket & 30\degree & 8/10 &  \bf 10/10 \\
 Sink handle & 30\degree & \bf 4/10 & 3/10 \\
 \addlinespace
 Handrail bracket & 60\degree & 1/10 & \bf 9/10 \\ 
 Ornamental handrail bracket & 60\degree & 2/10 & \bf 7/10 \\
 Sink handle & 60\degree & 0/10 & \bf 7/10 \\
 \bottomrule
\end{tabular}
\caption{\textbf{Physical Experiments Results for Convex Cavities: }We report the number of successful kitting trials for Kit-Net and the 2D baseline over 10 trials for 3 previously unseen objects with initial rotations of 30\degree~ and 60\degree~. Results suggest that Kit-Net significantly outperforms the 2D baseline for initial rotations of 60\degree~ and outperforms the baseline for two out of three objects for initial rotations of 30\degree~.}
\label{table:positive-results}
\end{table}
\subsection{Physical Kitting into Concave Conformal Cavities}
\label{subsubsec:real-negative}
\begin{figure}[t]
  \vspace{8pt}
  \centering
  \includegraphics[width=0.48\textwidth]{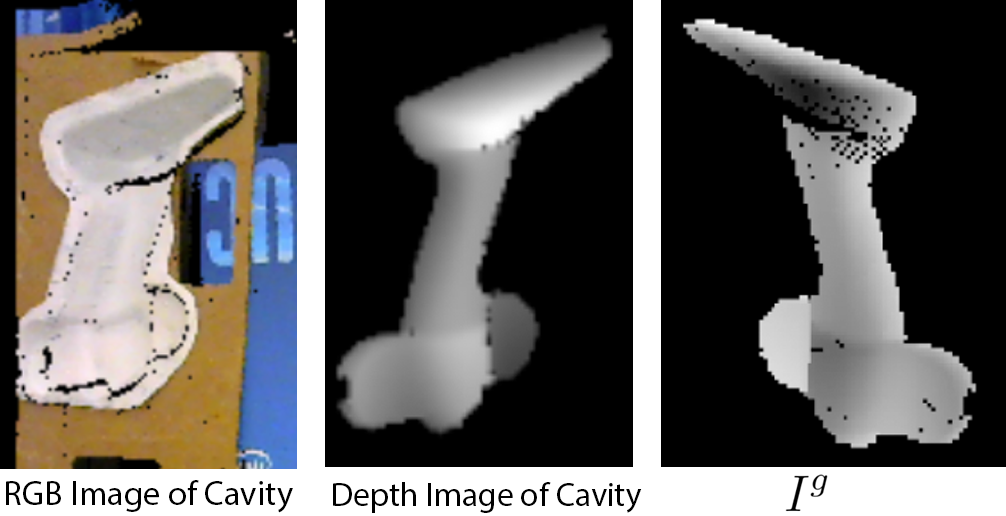}
  \caption{\textbf{Generating Negative Goal Images: }After taking the original image (left), we segment out all parts that do not belong to the cavity (middle). Then, we project from depth to point cloud, rotate the point cloud 180\degree~about the x axis centered at its centroid, and deproject to depth image to get $I^g$ (right).}
  \label{fig:rotated-cavity}
\end{figure}

The positive cavity task provides a detailed depth image, but requires the cavity to first be presented at a flipped, 180\degree~rotation for imaging and then flipped back for the kitting task, which may not be feasible in an industrial environment. 
Here, we perform the same experiment as in Section~\ref{subsubsec:real-positive}, but instead generate $I^g$ directly from an image of the cavity without flipping. Precisely, we segment out the cavity from an overhead depth image. Because we know the location of the cavity when setting up the kitting task, we can segment out the background and isolate the cavity. See Figure~\ref{fig:splash} (Left) for visuals. In application, a dedicated area for the cavities with a flat background would aid in the segmentation. Then, we deproject the segmented depth image into a point-cloud representation, rotate the point cloud 180\degree~around its center of mass, and project the rotated point cloud to the depth image $I^g$. This process is illustrated in Fig.~\ref{fig:rotated-cavity}.

\begin{table}[h]
\centering
 \begin{tabular}{l c c c}\toprule
 Object & Angle & 2D Baseline & Kit-Net \\
 \midrule
 Handrail bracket & 30\degree & 0/10 & \bf 9/10 \\ 
 Ornamental handrail bracket & 30\degree & 0/10 & \bf 7/10 \\
 Sink handle & 30\degree & 1/10 & \bf 3/10 \\
 \addlinespace
 Handrail bracket & 60\degree & 0/10 & \bf 7/10 \\ 
 Ornamental handrail bracket & 60\degree & 0/10 & 0/10 \\
 Sink handle & 60\degree & 1/10 & \bf 4/10 \\
 \bottomrule
\end{tabular}
\caption{\textbf{Physical Experiments Results for Concave Cavities: }We report the number of successful kitting trials for Kit-Net and the 2D baseline over 10 trials for 3 previously unseen objects with initial rotations of 30\degree~ and 60\degree~. Results suggest that Kit-Net significantly outperforms the baseline in all settings except for the handrail bracket with an initial rotation of 60\degree~, for which neither Kit-Net nor the baseline can successfully kit the object.}
\label{table:negative-results}
\end{table}

\begin{figure}[ht]
  \vspace{8pt}
  \centering
  \includegraphics[width=0.48\textwidth]{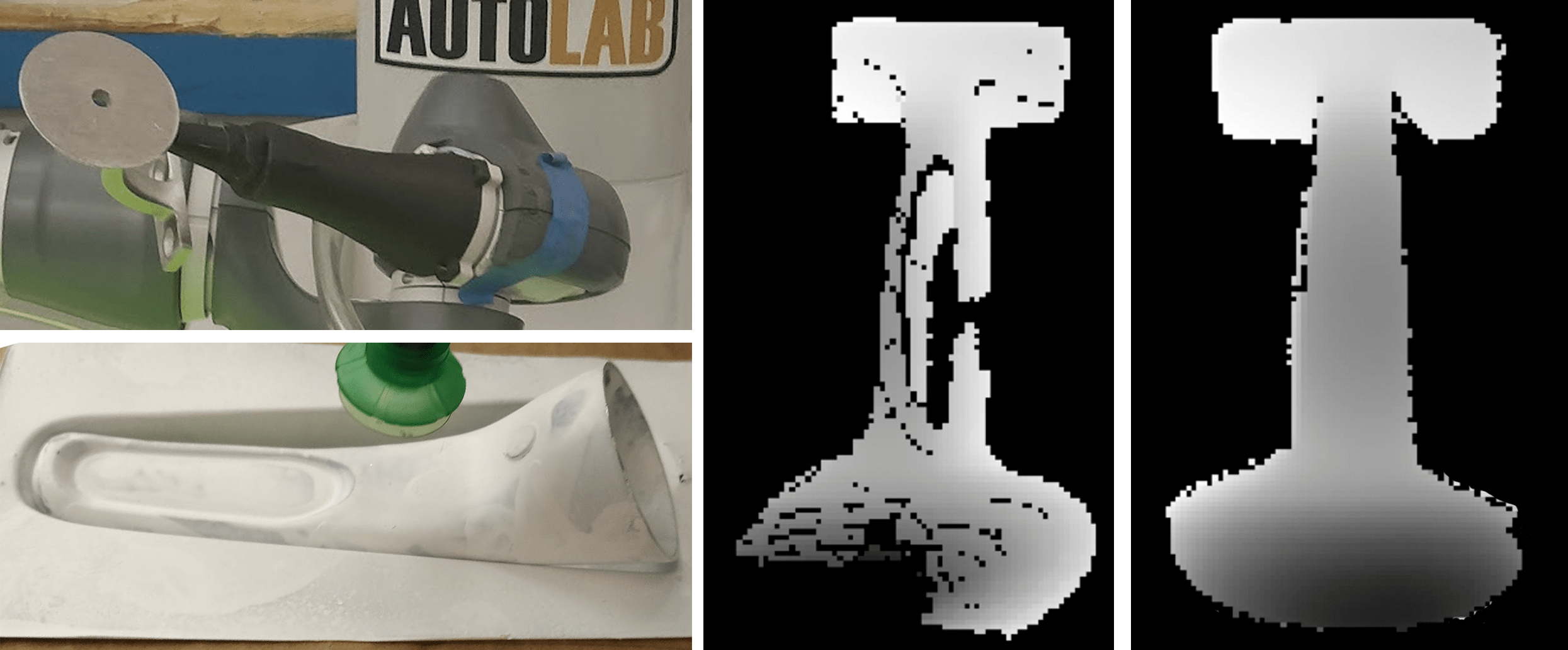}
  \caption{\textbf{Kit-Net Failure Cases: } The top-left image shows a configuration of the ornamental handrail bracket
  where the suction gripper occludes
  the handle below the base. The bottom-left image shows the sink handle. Although Kit-Net was able to orient the handle correctly for insertion, the centroid matching had a small error in estimating translation and the cavity does not have enough slack to be properly inserted. The center and right images show depth images for the ornamental handrail bracket for the concave conformal cavity and convex conformal cavity, respectively.
  The inside of the concave cavity is very thin and the angle of the camera makes it hard to perfectly image it, resulting in a poor depth image (center image). 
  This contributes to the failures for both the baseline and for Kit-Net when the initial rotation is 60\degree{} away from the desired rotation for kitting.}
  \label{fig:failure-modes}
\end{figure}

Table~\ref{table:negative-results} shows results from experiments with 3 novel objects from Fig.~\ref{fig:clamshell-objects} across 10 controller rollouts. We observe that Kit-Net outperforms the baseline for initial rotations of both 30\degree~ and 60\degree~ on the handrail bracket and sink handle, and for an initial rotation of 30\degree~ for the ornamental handrail bracket. For the ornamental handrail bracket, the depth image from the concave cavity is low quality as shown in Fig.~\ref{fig:failure-modes} (center image), causing Kit-Net to fail when the object is 60\degree~ away the correct insertion orientation. We examined this failure and found that it occurs because the cavity for the neck of the bracket is very thin, making it difficult to obtain a good depth image. Kit-Net also has low performance on the sink handle due to small errors in centroid matching, as discussed in Section~\ref{subsubsec:real-positive}. Fig.~\ref{fig:failure-modes} (bottom left) shows an example failure case where the sink handle is correctly oriented but the translation is slightly off. Fig.~\ref{fig:failure-modes} (top left) shows an example of an occasional cases where the suction gripper occludes the handle of the ornamental handrail bracket. In these cases, the robot can only see the base, resulting in failure.

\section{Discussion and Future Work}
\label{sec:discussion}
We present Kit-Net, a framework that uses self-supervised deep learning in simulation to kit novel 3D objects into novel 3D cavities. Results in simulation experiments suggest that Kit-Net can kit unseen objects with unknown geometries into a prismatic target in less than 5 controller steps with a median percent fit of 99\,\%.
In physical experiments kitting novel 3D objects into novel 3D cavities, Kit-Net is able to successfully kit novel objects 63\,\% of the time while a 2D baseline that only considers $SE(2)$ transforms only succeeds 18\,\% of the time.
In future work, we will work to improve performance by using the predicted error from Kit-Net
to regrasp the object in a new stable pose~\cite{tournassoud1987regrasping,dafle2014extrinsic,danielczuk2020exploratory} before reattempting the kitting task, study Kit-Net's performance with other depth sensors, and apply Kit-Net to kit objects that are initially grasped from a heap \cite{murali20206,danielczuk2020x}.
\section{Acknowledgments}
\begin{scriptsize}
\noindent This research was performed at the AUTOLAB at UC Berkeley in affiliation with the Berkeley AI Research (BAIR) Lab, Berkeley Deep Drive (BDD), the Real-Time Intelligent Secure Execution (RISE) Lab, and the CITRIS ``People and Robots" (CPAR) Initiative. Authors were also supported by the Scalable Collaborative Human-Robot Learning (SCHooL) Project, a NSF National Robotics Initiative Award 1734633, and in part by donations from Siemens, Google, Toyota Research Institute, Autodesk, Honda, Intel, Hewlett-Packard and by equipment grants from PhotoNeo and Nvidia. This material is based upon work supported by the National Science Foundation Graduate Research Fellowship Program under Grant No. DGE 1752814. Any opinions, findings, and conclusions or recommendations expressed in this material are those of the authors and do not necessarily reflect the views of the sponsors.
\par 
\end{scriptsize}


\renewcommand*{\bibfont}{\footnotesize}
\printbibliography 

\end{document}